\documentclass[runningheads]{llncs}
\usepackage{graphicx}
\usepackage{bbm}

\usepackage[colorlinks]{hyperref}

\usepackage{tabularx}
\usepackage{booktabs,multirow}
\usepackage[acronym,shortcuts,nonumberlist]{glossaries}
\glsdisablehyper

\newacronym{rdc}{RDCNet}{recurrent dilated convolutional network}
\newacronym{sdc}{SDC}{stacked dilated convolution}
\newacronym{ssdc}{sSDC}{shared stacked dilated convolution}
\newacronym{esj}{ESJ}{embedding soft jaccard}
\newacronym{sbd}{SBD}{symetric best dice}
\newacronym{aji}{AJI}{aggregated Jaccard index}

\newcommand{\figref}[1]{Fig.~\ref{#1}}
\newcommand{\secref}[1]{Sec.~\ref{#1}}
\newcommand{\tableref}[1]{Table~\ref{#1}}

\newcommand{\ie}{\textit{i.e.\ }}
\newcommand{\eg}{\textit{e.g.\ }}

\newcommand{\inputvar}{X}
\newcommand{\outputvar}{Y}
\newcommand{\embeddingvar}{y}
\newcommand{\scalingvar}{s_d}
\newcommand{\marginvar}{margin}

\begin{document}
\title{RDCNet: Instance segmentation with a minimalist recurrent
  residual network} 

\titlerunning{Instance segmentation with a
minimalist recurrent residual network} 

\author{
  Raphael Ortiz\inst{1}\and
  Gustavo de Medeiros\inst{1}\and
  Antoine H.F.M. Peters\inst{1,2}\and
  Prisca~Liberali\inst{1,2}\and 
  Markus Rempfler\inst{1}}

\authorrunning{R.~Ortiz et al.}

\institute{
 Friedrich Miescher Institute for Biomedical Research (FMI), Basel, Switzerland \and
 Faculty of Sciences, University of Basel, Basel, Switzerland
}

\maketitle

\begin{abstract}
  Instance segmentation is a key step for quantitative microscopy. 
  While several machine learning based methods have been proposed for 
  this problem, most of them rely on computationally complex models 
  that are trained on surrogate tasks. Building on recent developments 
  towards end-to-end trainable instance segmentation, we propose a 
  minimalist recurrent network called \acf{rdc}, consisting of a 
  \acf{ssdc} layer that iteratively refines its output and thereby 
  generates interpretable intermediate predictions. It is light-weight 
  and has few critical hyperparameters, which can be related to 
  physical aspects such as object size or density.We perform a 
  sensitivity analysis of its main parameters and we demonstrate its 
  versatility on 3 tasks with different imaging modalities: nuclear 
  segmentation of H\&{}E slides, of 3D anisotropic stacks from 
  light-sheet fluorescence microscopy and leaf segmentation of top-view 
  images of plants. It achieves state-of-the-art on 2 of the 3 
  datasets.

\end{abstract}

\section{Introduction}

\begin{figure}[t]
  \includegraphics[width=\textwidth]{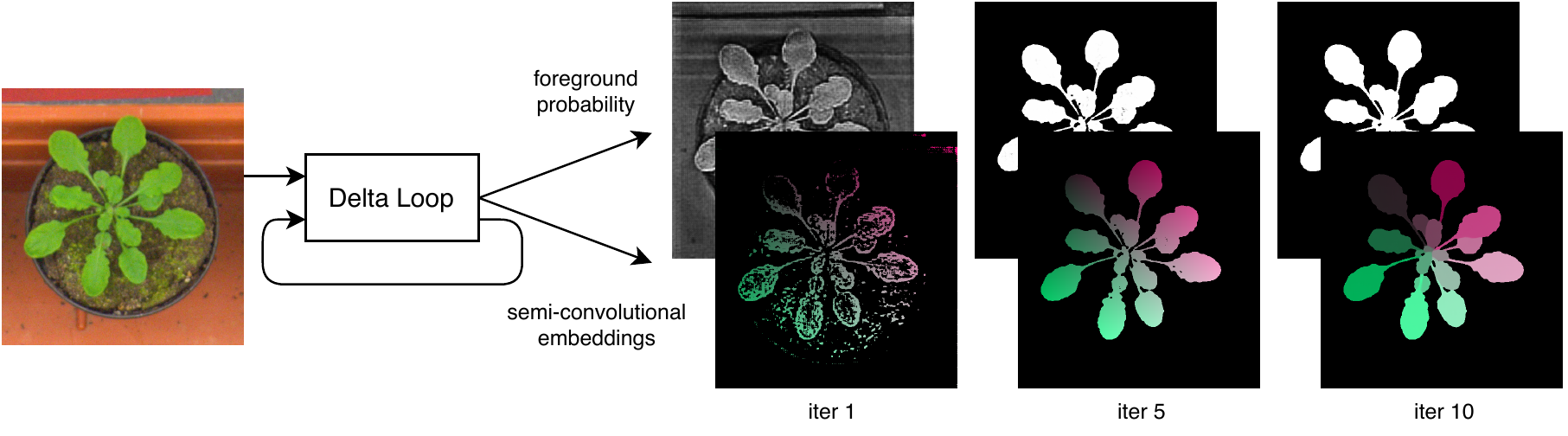}
  \caption{Our approach to instance segmentation illustrated on an
    example of leaf segmentation. A minimalistic network is applied
    recursively to the input, refining its predictions of foreground
    and instance embeddings iteratively. Foreground probabilities are
    depicted in greyscale (white $\simeq P(\mathrm{foreground})=1$),
    embeddings are shown in pseudocolor. While the outputs of the
    first iteration $i=1$ are not very coherent yet, they quickly
    converge until, in the last iteration $i=10$, the embeddings
    sharply distinguish even the slim stalk of the individual leafs.}
  \label{fig:teaser}
\end{figure}

Instance segmentation is the task of detecting and outlining different objects of the same class within an image, \eg cells or nuclei. As such, it is a key step in
many biomedical image analysis workflows and has recently received
increased attention in the
community~\cite{Chen2019,de2017,Kumar2019,payer2018,Payer2019,Schmidt2018,Weigert2020}.

Instance segmentation was traditionally tackled with pipelines based on watershedding, while more recent approaches rely mostly on machine learning. However due to the permutation invariance of instance labels, training machine-learning based models end-to-end is non-trivial, and remains an active area of research. A first class of methods are proposal-driven and follow a detect, then segment paradigm. For instance, Mask-RCNN \cite{he2017} first predict object bounding boxes and then perform mask segmentation within each box. StarDist~\cite{Schmidt2018,Weigert2020}, removes this last step by predicting a star-convex polyhedron instead of rectangular boxes. This is often more robust, but the fact that a fixed shape representation approximates the object, limits the accuracy of the segmentation boundary and restricts its application to roundish shapes. 

The second class of methods tackles the problem in reverse, segment first, then detect. Early approaches were often trained on surrogate outputs such as foreground vs background vs boundary, which are then combined during post-processing, \ie not end-to-end~\cite{Ronneberger2015,Kumar2019}. More recently, \emph{Instance embedding} methods~\cite{Chen2019,de2017,Neven2019,payer2018,Payer2019} aiming to predict pixel embeddings that are similar within an instance and dissimilar across different instances have been proposed. However, losses like the ones used~\cite{Chen2019,de2017,payer2018,Payer2019} are also surrogates. Only lately, Neven et al.~\cite{Neven2019} proposed to optimize a segmentation loss to learn the instance embedding.

As argued in~\cite{Novotny2018}, embedding methods
like~\cite{Chen2019,de2017,payer2018,Payer2019} that rely on fully convolutional models
have to rely on texture to generate embeddings that satisfy the
(dis-)similarity requirements because of their translation
invariance. In turn, this makes the task of modelling the embedding
considerably more complex and, for example, creates problems when
having to process an input image in a tiled fashion.

Besides specifying an appropriate loss to optimize for the given task, the network architecture also plays a crucial role. U-Net-like~\cite{Ronneberger2015} architectures are at the core of many state-of-the-art biomedical segmentation methods. While increasingly complex architectures are being proposed, state-of-the-art results in segmentation tasks have recently been demonstrated with a more light-weight U-Net applied recurrently to refine its output \cite{Wang2019}.

In this paper, we take it a step further and propose a minimalist
\ac{rdc}
for instance segmentation in biomedical instance segmentation. We
distill knowledge from a few recent key works on instance embeddings,
namely 1) to use semi-convolutional~\cite{Novotny2018} or its special
case, the additive semi-convolutional layer to generate the embeddings,
and 2) train the embedding output on an instance segmentation loss
similar to~\cite{Neven2019}. Combining those, the instance
segmentation task becomes much simpler and we propose 3) to address it
with a light-weight, recurrent architecture which is applied
iteratively to input and its latest prediction. This minimalist
\ac{rdc}
has only a fraction of the parameters of a common
U-Net~\cite{Ronneberger2015} and is much easier to interpret since the
output at every iteration can easily be visualized (see
\figref{fig:teaser}).

Finally, we conduct experiments on three different datasets and show
that our approaches matches state-of-the-art without extra
domain-specific adjustments.

\section{Methods}

\begin{figure}[t]
  \includegraphics[width=\textwidth]{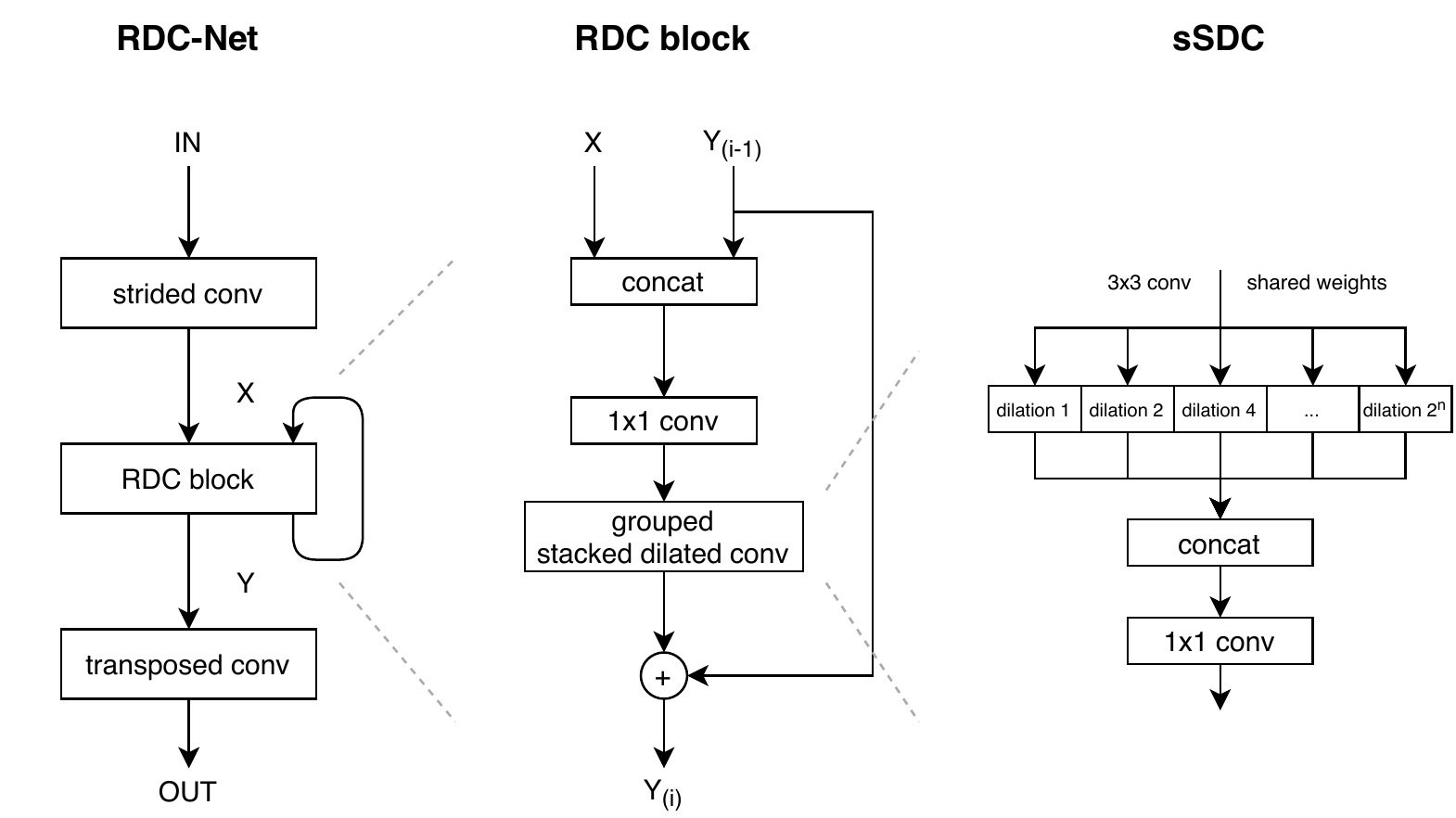}
  \caption{Our overall network architecture (\emph{left}), the
    recursive module (\emph{center}) that refines a previous
    prediction $Y^{(i-1)}$ for $X$ in a residual fashion, and
    (\emph{right}) its core component, a stacked, dilated convolution
    block with shared weights (\acs{ssdc}) followed by a $1 \times 1$
    convolution to project the stacked features back to the same
    number of channels as the input.}
  \label{fig:architecture}
\end{figure}

\subsubsection{Architecture rational.} Our architecture consists of a single shallow recurrent block sandwiched between learnable downsampling and upsampling layers respectively as depicted in \figref{fig:architecture} (left). In the 3D case, these layers have a different stride in XY than in Z to compensate for anisotropy of typical microscopic Z-stacks. At the heart of this (\figref{fig:architecture}, center) is a recurrent block that follows a residual formulation and is comprised of grouped convolutions of kernel size 3. A point-wise, \ie $1\times 1$, convolution is also added to mix group information between iterations. To bring global context in early iterations, each group uses stacked dilated convolutions \cite{Schuster2019} with shared weights and a point-wise convolution for channel reduction, hereafter called \ac{ssdc} and depicted in \figref{fig:architecture} (right). Note that this procedure enables the network to determine when to use narrow/wide convolutions based on training data. All convolutional layers except the first one are preceded by leaky ReLU ``pre-activation''  to keep the residual path clear \cite{he2016}.

Formally, our \ac{rdc} generates the output of iteration $i$
based on input $\inputvar$ and its previous output $\outputvar^{i-1}$
\begin{equation}
  \label{eq:recursion}
  \outputvar^{i} = f_{\theta}(\inputvar, \outputvar^{i-1}) + \outputvar^{i-1} \enspace,
\end{equation}
where $f_{\theta}$ is the transform of the recurrent block and $\outputvar^0$ is initialized to zero.

The raw network output is split in two branches. The first predicts the
semantic class with a softmax activation, \ie in this work simply
foreground-background. The other predicts the instance embeddings and
is chosen to be an additive semi-convolutional
layer~\cite{Novotny2018} that sums pixel/voxel coordinates to the
 convolutional output, leading to semi-convolutional
instance embeddings in 2D or 3D, respectively. An illustration of the semi-convolutional layer is provided in supplement, Fig.~1.

\subsubsection{Objective functions.} At training time, we follow~\cite{Neven2019} and convert embeddings
$\embeddingvar$ into probabilities for pixel $u$ being part of
instance $k$ as
\begin{equation}
  \label{eq:esj}
  P\left(u=k \right) = \exp \big(-\frac{ \Vert \embeddingvar_u - \hat{\embeddingvar}_k \Vert^2}{2 \sigma^2}\big) \enspace,
\end{equation}
where $\sigma > 0$ is the bandwidth parameter and the centroids
$\hat{\embeddingvar}_k$ are estimated as the mean embedding
under the \emph{true} mask of instance $k$ denoted with
$\mathcal{S}_k$, \ie
  $\hat{y}_k = \frac{1}{\vert \mathcal{S}_k\vert} \sum_{u \in \mathcal{S}_k} y_u$.
Note that the bandwidth $\sigma$ can be reformulated as a more interpretable $\marginvar$ parameter defined by the distance from the centroid where $P\left(u=k \right) = 0.5$:
\begin{equation}
  \label{eq:margin}
  \marginvar = \sigma \sqrt{-2 \ln{0.5}} \enspace .
\end{equation}
Ultimately, this generates a probability map for each instance such that
it can be directly compared to the one-hot encoded \emph{true}
segmentation $\mathbbm{1}\lbrace{u=k\rbrace}$. 
In our workflow, we choose to optimize the soft Jaccard loss as defined in~\cite{Rahman2016} for both class and instance predictions (\ie an \acf{esj}) as it inherently handles class imbalance. Using the same loss function for both tasks also facilitates multi-task training such that we found uniform loss weighting to be sufficient. Finally, since we only use the instance embeddings to split the predicted foreground mask into instances, the embedding loss is not applied over background regions.

\subsubsection{Post-processing.} At inference time, the \emph{true} centroids
used in \eqref{eq:esj} are not available. Hence, we have to estimate
them from the embeddings. We do this by a computationally inexpensive
Hough voting scheme as illustrated in supplement, Fig.~2. Embeddings are binned in a 2D/3D histogram where
each bin corresponds to a pixel/voxel in the input image. Local maxima
with a window size related to $\marginvar$ are taken as centres. With a perfect zero loss, the inter-center distance would be at least $2~\marginvar$, however in practice we find it beneficial to fine tune the window size on the validation split. Finally, instance labels are obtained by assigning foreground pixels
to their nearest center in the embedding space.

\section{Experiments \& Results}

\begin{figure}[t]
  \includegraphics[width=\textwidth]{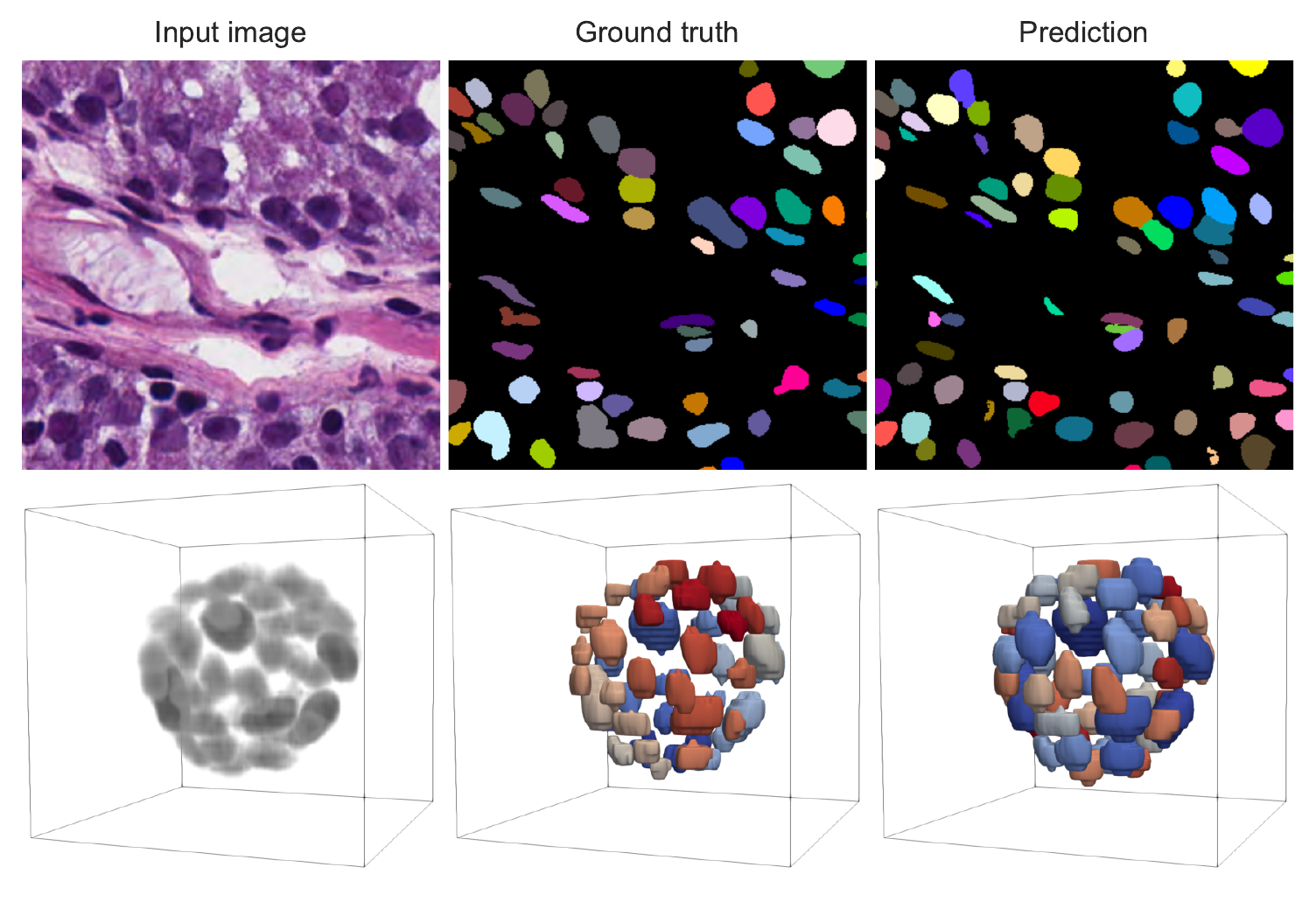}
  \caption{Qualitative results of our \acs{rdc} on images from
    MoNuSeg (top row) and 3D-ORG (bottom row). For 3D-ORG,
    \emph{unlabeled} pixels at the object boundary are not depicted
    in the ground truth and therefore appear slightly smaller than the
    predictions.}
  \label{fig:qualitative}
\end{figure}

\begin{figure}
  \includegraphics[width=\textwidth]{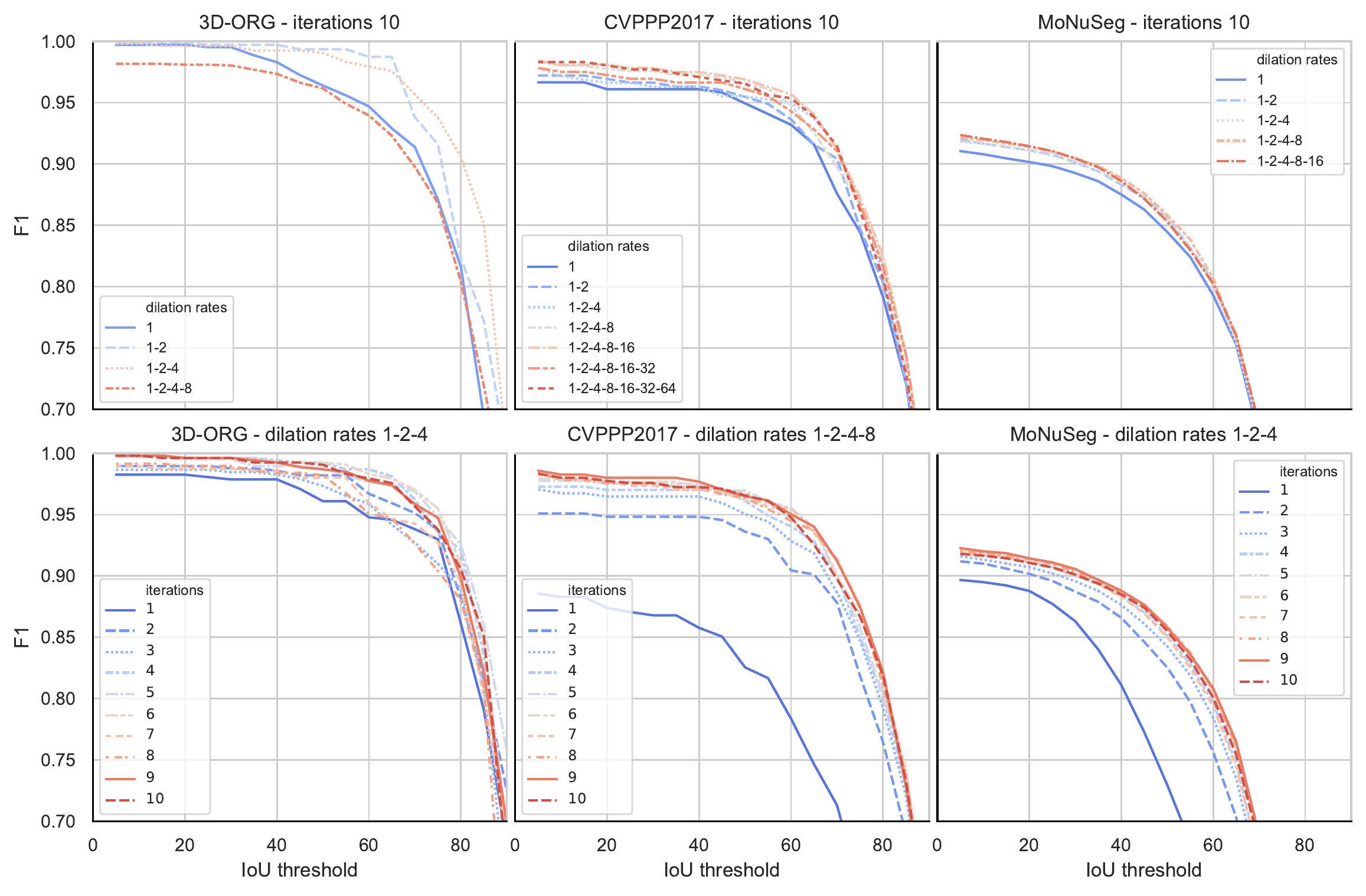}
  \caption{Hyperparameter analysis. We compare the effect of different
    dilation ranges in the \ac{ssdc} block (top) and varying number of
    iterations (bottom) on all three datasets (columns). Curves depict instance F1-score with varying IoU threshold. 
    Dilation rate=1 corresponds to the base case of regular 3x3 convolution.
  }
  \label{fig:analysis}
\end{figure}

\begin{table}
  \caption{ Quantitative evaluation of instance
    segmentations. Precision, recall and F1 scores are calculated on a
    per-instance basis at an intersection-over-union acceptance
    threshold of 0.5. \Acf{sbd} and \acf{aji} are calculated as in the
    respective challenges \cite{Minervini2016} and \cite{Kumar2019}.
    Scores for competing methods on public datasets are taken from their respective publication.}
  \label{tbl:scores}
{
\newcolumntype{b}{X}
\newcolumntype{m}{>{\hsize=.55\hsize}X}
\newcolumntype{s}{>{\hsize=.30\hsize\raggedleft\arraybackslash}X}
\begin{tabularx}{\textwidth}{mbsssss}
  \toprule
  Dataset & Method               &  Precision &     Recall &         F1 &       SBD &       AJI \\
  \midrule
  \multirow{5}{*}{CVPPP2017}
  & Hglass + cos similarity~\cite{payer2018}   &     - &      - &      - &      84.5 &      - \\
  & Recurrent attention~\cite{Ren2017}   &     - &      - &      - &      84.9 &      - \\
  & Mask-RCNN~+~synth~\cite{Ward2019}   &      - &      - &      - &   90.0 &      - \\
  & UNet + semi-conv + \ac{esj}                  &     96.9 &      94.1 &      95.4 &      89.6 &      89.9 \\
  & Ours                        &     97.3 &      96.7 &      96.9 &     \textbf{91.4} &      96.9 \\
  \midrule \multirow{5}{*}{MoNuSeg}
  & Dual UNet \cite{Li2019}       &      - &      - &      79.1 &      - &      59.0 \\  
  & ResNet + bending loss \cite{wang2020bending}       &      - &      - &      77.2 &      - &      63.0 \\
  & MoNuSeg top score \cite{Kumar2019}       &      - &      - &      - &      - &      \textbf{69.1} \\
  & MoNuSeg inter-human \cite{Kumar2019}       &      - &      - &      - &      - &      65.3 \\
  & UNet + semi-conv + \ac{esj}                  &     79.8 &      90.8 &      84.8 &      72.7 &      66.1 \\
  & Ours                        &     82.0 &      90.3 &      85.8 &      74.2 &      67.3 \\
  \midrule \multirow{3}{*}{3D-ORG}
  & StarDist~\cite{Weigert2020} &     97.5 &      95.7 &      96.6 &      87.7 &      80.2 \\
  & UNet + semi-conv + \ac{esj} 				  &     98.1 &      94.6 &      96.2 &      92.1 &      86.2 \\
  & Ours                        &     \textbf{99.4} &      \textbf{99.1} &      \textbf{99.3} &      \textbf{95.5} &      \textbf{92.4} \\
  \bottomrule
\end{tabularx}
}
\end{table}

We conduct experiments on three datasets of different modalities and compare our results to the literature where available as detailed below. 
For all experiments, we train a \ac{rdc} with 8 parallel \ac{ssdc} blocks, \ie groups, each having 64 channels. The number of iterations is varied from 1 to 10 and the set of dilation rates of the \ac{ssdc} block from 1 up to the largest that can fit in the training patch size. A scaling factor $\scalingvar$ is chosen according to the instances size and image resolution of each dataset $d$ as detailed below. The strided convolution input layer has both kernel size and stride of $\scalingvar$ and 32 channels while the output transposed convolution has a kernel size of $2\scalingvar$ and stride $\scalingvar$. We train all networks with Adam optimizer and cosine scheduler with learning rates from $10^{-3}$ to $10^{-5}$, 10\% spatial dropout on the input of the recurrent block and a batch size of 2. We also perform online data augmentation as detailed in supporting information.
To assess the benefits of our architecture, we also train a U-Net~\cite{Ronneberger2015} baseline
using the same \ac{esj} loss and a additive semi-convolutional layer for the embedding.
We report the instance precision, recall and $F_1$ scores at 50\,\% IoU threshold per-image unless otherwise specified. In accordance with relevant literature, we also report the \ac{sbd} \cite{Minervini2016} and \ac{aji} \cite{Kumar2019}.

\subsection{Datasets}
\label{sec:datasets}

\paragraph{CVPPP2017:} A plant leaf segmentation dataset from
\cite{Cvppp2015,Minervini2016}. Leafs exhibit more complex shapes
than nuclei and are therefore an excellent test case for a general
instance segmentation method. An example can be found in
\figref{fig:teaser}. Since ground-truth for the test set is not publicly available, we randomly split the 128 images of the A1 subset into 60\,\% training, 20\,\% validation and 20\,\% test. We train on $480 \times 480$ crops and use a scaling factor of 4. The $\marginvar$ in the \ac{esj} is set to 10\,px. Note that this drives centers of close-by, central leaves apart while centers of mass of large leaves remain valid solutions. We compare our results with 3 methods reported in the literature: cosine embedding that uses local constraints \cite{payer2018}, recurrent attention that models a human-like counting process \cite{Ren2017} and a recent work based on Mask-RCNN with domain-tailored data augmentation \cite{Ward2019}.

\paragraph{MoNuSeg:} A nuclei segmentation dataset composed of digital
pathology images originating from seven different
organs~\cite{Kumar2019}. The training set contains 30 images that we re-split into 23 and 7 images for training and validation respectively. We report results on the 14 test images as described in the original challenge. We train on $256 \times 256$ crops, using a scaling factor of 4 and 5 px $\marginvar$ for the \ac{esj} loss. Due to the smaller field of view required for this dataset, the U-Net baseline is trained with 4 levels instead of the original 5 which provided better results. We compare our result to the MoNuSeg competition~\cite{Kumar2019} top rank as well as published work relying on boundary prediction \cite{Li2019} and a curvature regularization technique \cite{wang2020bending}.

\paragraph{3D-ORG:} A 3D nuclei segmentation dataset comprised
of 49 3D image stacks of developing organoids acquired with light-sheet
microscopy. The stacks have a voxel spacing of
$0.26 \times 0.26 \times 2~\mu\mathrm{m}$. An example is depicted in
\figref{fig:qualitative}.  We split the dataset into 32, 9 and 8 images for
training, validation and testing respectively. Annotations were obtained in a
semi-automatic fashion with manual corrections. Cell boundaries sometimes appear fuzzy due to low signal-to-noise  and scattering. In these ambiguous regions, the annotation labels were set as \emph{undefined}, which can be elegantly handled in our approach by zeroing the loss in these regions. We train on $20 \times 160 \times 160$ crops, using a scaling factor of (1,8,8), \ie no downscaling along Z to compensate for anisotropy. In this case the semi-convolutional layer also uses physical coordinates rather than voxel indices. The \ac{esj} loss $\marginvar$ is set to $3\,\mu\mathrm{m}$. Anisotropy in the U-Net baseline is compensated by omitting max-pooling along the Z-axis. During post-processing, we clean jagged edges with a label-wise morphological opening of $0.5~\mu\mathrm{m}$. Finally, we compare our method against StarDist~\cite{Weigert2020}. Since StarDist relies on explicit boundaries for training, voxels in undefined regions are assigned to the nearest label.

\subsection{Hyperparameter analysis}

We examine the behavior of our \ac{rdc} with varying hyperparameters.
To this end, we vary the dilation rates from 1 (\ie none) up to 64, depending on the dataset, and the number of iterations in the range $i \in \lbrace 1, \ldots 10\rbrace$. The results are depicted in \figref{fig:analysis}.

We observe that for a fixed number of iterations $i=10$, the F1 score increases with higher dilation rates. Since the weights of the spatial convolutions are shared, the network complexity is only slightly increased due to the subsequent point-wise convolution. Therefore, improvement is likely due to the larger receptive field available during early iterations. This is in line with observations that several small stacked hourglass networks perform better than a large one as reported in \cite{Newell2016}. One exception we notice is on the 3D-ORG dataset, where increasing dilations up to 8 decreases F1. We account this to a field of view that is too big for the available patch size.

Looking at the effect of the number of iterations (\figref{fig:analysis} bottom row), we note that the increased context from dilations alone (\ie only one iteration) is not sufficient. Clear improvements are observed with increasing number of iterations, plateauing at around 5 iterations for the three datasets.
This indicates that 5 iterations would be a good choice to balance performance with computational complexity, resulting in  68\% Multiply-Add operations of the baseline UNet.

\subsection{Comparison}
Quantitative evaluation results on all 3 datasets
(described in \secref{sec:datasets})  are shown in \tableref{tbl:scores}. We
report state-of-the-art results on CVPPP2017 and 3D-ORG in terms of
\ac{sbd} and \ac{aji}, respectively. On MoNuSeg, our result falls
between rank 5 and 6 reported in~\cite{Kumar2019} without any extra
domain-specific modelling.  Note that this is already better than the
reported inter-human agreement level. With all 3 datasets we observe
an improvement across all metrics when swapping the baseline U-Net for
our \ac{rdc}, despite having $\approx 30 \times$ less parameters.
Importantly, it has a reduced memory footprint, which is particularly useful for processing 3D datasets. For example, predicting a single 3D patch of $32\times256\times256$ with RDCNet uses only 3.8\,GB VRAM while the baseline UNet uses 8.8\,GB. Furthermore, the memory footprint at prediction is constant w.r.t. the number of iterations.

\section{Conclusions}

We have presented a minimalist recursive network for instance
segmentation in biomedical images called \ac{rdc}.
We have shown that it achieves state-of-the-art results on 2 out of 3
datasets and is highly competitive on the other. This is achieved
by optimizing an instance segmentation loss instead of a surrogate and by relying on
semi-convolutional embeddings. We also advocated for grounding
hyperparameters in physical space which is made possible by the tight
relationship between embeddings and the spatial domain and provided
rules of thumb to adapt the method to different modalities. The
iterative nature of the \ac{rdc} facilitates the interpretation of
intermediate predictions, while the light-weight aspect of the
architecture makes it an ideal candidate to process volumetric
data. This could, potentially, even incorporate the temporal
dimension, which we plan to investigate in future work.

\bibliographystyle{splncs04}
\bibliography{references}

\end{document}


\maketitle
\appendix
\section{Data augmentation}
\begin{table}
  \caption{Data augmentation parameters used for experiments with the
    different datasets. Note that empty fields indicate that the
    particular augmentation method was not used for that dataset.}
  \label{tbl:augmentation}
{
  \newcolumntype{b}{X}
  \newcolumntype{m}{>{\hsize=.60\hsize\raggedright\arraybackslash}X}
\newcolumntype{s}{>{\hsize=.40\hsize\raggedleft\arraybackslash}X}
\begin{tabularx}{\textwidth}{bmssss}
  \toprule
  &                 &  CVPPP2017 & MoNuSeg & 3D-ORG \\
  Augmentation      & Parameter &&&\\
  \midrule \multirow{2}{*}{Random axis flip}
  & $p_{\mathrm{flip}}$  & 0.5       & 0.5      & 0.5    \\
  & Axes               & X, Y       & X, Y    & X, Y, Z   \\
  \midrule
  \multirow{2}{*}{\shortstack{Random offset $\sim \mathcal{N}(\mu, \sigma)$ \\ \tiny{(IID per patch)}}}
  & $\mu$              & 0          & 0        & 0      \\
  & $\sigma$           & 0.2        & 0.2      & 0.2    \\
  \midrule
  \multirow{2}{*}{\shortstack{Random noise $\sim \mathcal{N}(\mu, \sigma)$ \\ \tiny{(IID per pixel)}}}
  & $\mu$              &            &          & 0.05   \\
  & $\sigma$           &            &          & 0.3    \\
  \midrule
  \multirow{3}{*}{Random HSV shift}
  & $\Delta$ Hue       & $\pm$ 0.3     & $\pm$ 0.3  & \\
  & Saturation         & $ [0.8, 1.2]$ & $[0.8, 1.2]$ & \\
  & Value              & $ [0.8, 1.2]$ & $[0.8, 1.2]$ & \\
  \midrule
  \multirow{3}{*}{\shortstack{Random Gaussian blur\\ $\sigma$ range}}
  & $p_{\mathrm{active}}$ & 0.5         & 0.5 &\\
  & $\sigma$            & [0.5, 3]           & [0.5, 2] &\\
  \midrule
  \multirow{3}{*}{Random affine transform}
  & zoom factor        & [0.9, 1.1] & [0.9, 1.1]  &\\
  & shear              & $\pm 5$    & $\pm 5$  &\\
  & rotation angle     & $\pm 10$   & $\pm 10$  &\\
  \midrule
  \multirow{1}{*}{Random warp}
  & A      & 20         & 20 & \\
  \midrule
  \multirow{4}{*}{Random clipping $\sim \mathcal{N}(\mu, \sigma)$}
  & \textbf{$\mu_{min}$}      & -1         & -1 &\\
  & \textbf{$\mu_{max}$}      & 1         & 1 &\\
  & \textbf{$\sigma$}      & 0.3        & 0.3&\\
  \bottomrule
\end{tabularx}
}
\end{table}

We detail the data augmentation used during training for all our
experiments in Table~\ref{tbl:augmentation}. Random warp was generated by smoothing uniformly distributed pixel offset in range [$-A$, $A$] with a gaussian kernel $\sigma = 2~A$ where $A$ is the max amplitude.

\section{Semi-convolutional embedding}
\begin{figure}[H]
  \includegraphics[width=0.7\textwidth]{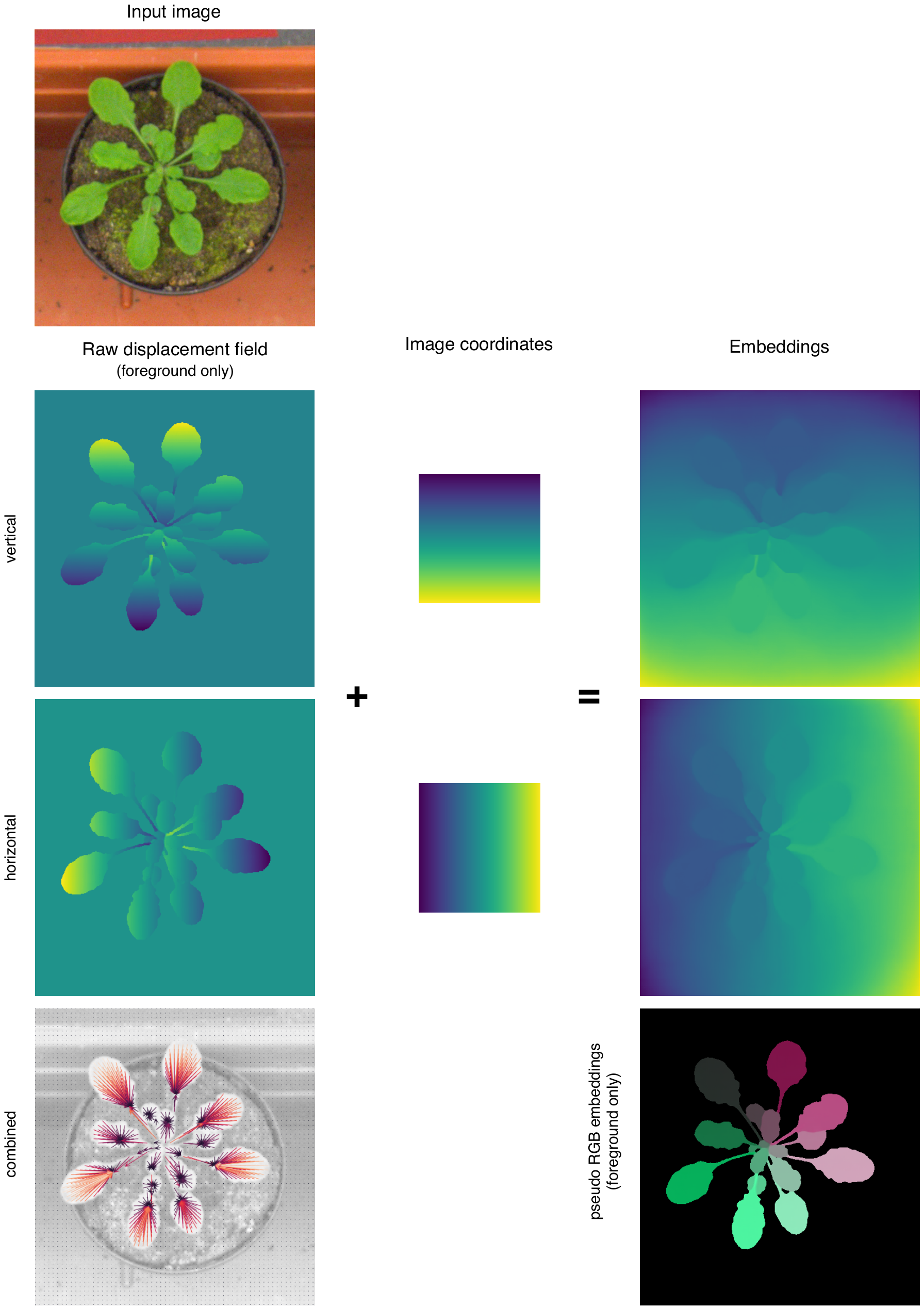}
  \caption{Example of how so called semi-convolutional embeddings \cite{Novotny2018} are formed. Given the raw image on the top left, the network output a displacement field where all pixels of a given instance point to the same location. Note that this centre of attraction is not restricted to the center of mass or any specific point of the object. Adding the image coordinates at each position produces the final embeddings where each instance is $\approx$assigned a unique embedding. This works regardless of the size of the image as the effective field of the network does not need to cover the entire image.}
  \label{fig:semiconv}
\end{figure}
\newpage

\section{Centroids estimation during inference}
\begin{figure}[H]
  \includegraphics[width=0.8\textwidth]{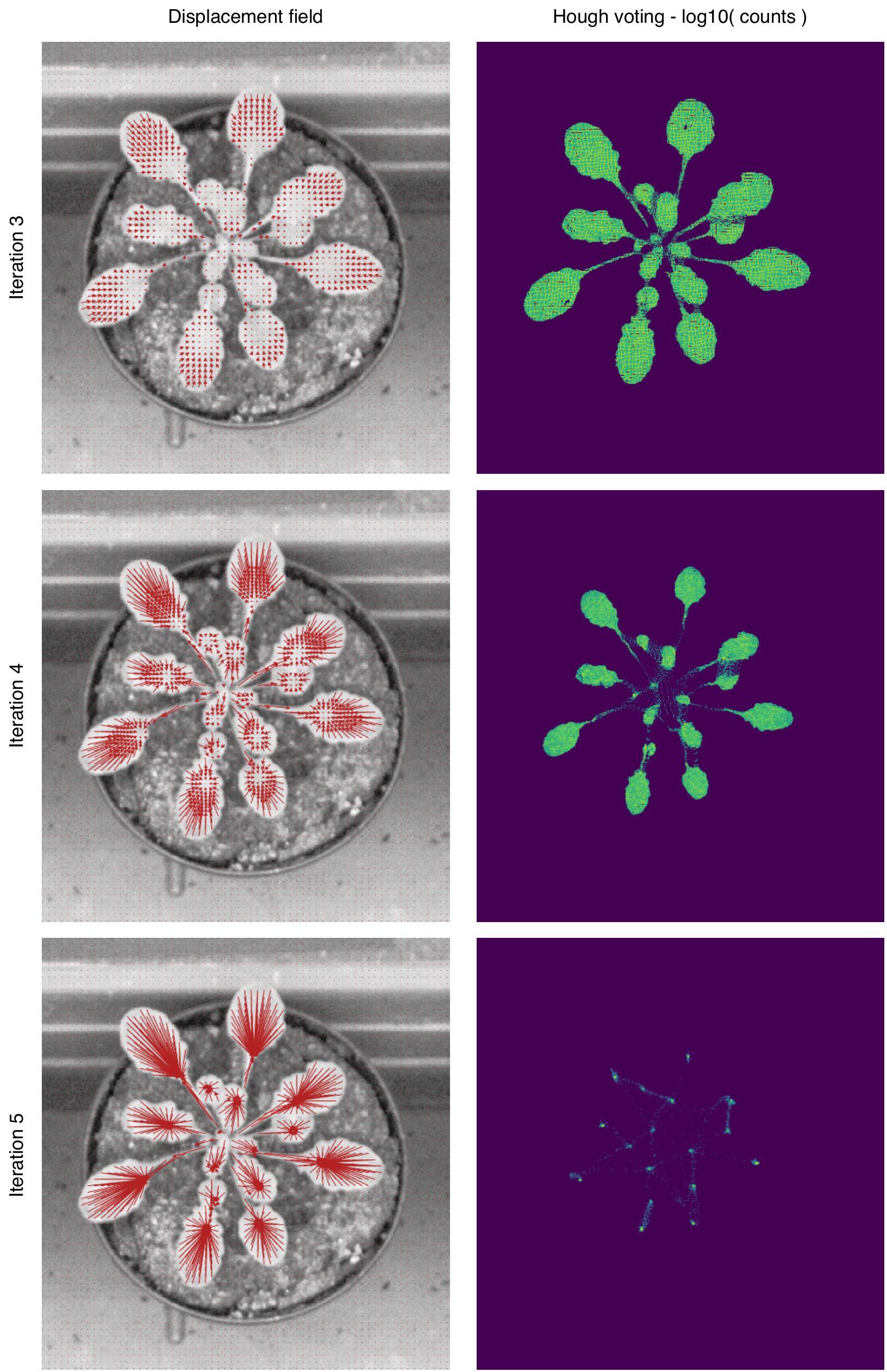}
  \caption{Example illustrating post-processing of the network output to obtain the centres used for final embedding assignment. The model was trained with 5 iterations. Initially, each pixel is voting for himself and instances are gradually separated with each iteration until they can be identified with a trivial Hough voting step.}
  \label{fig:semiconv}
\end{figure}

\bibliographystyle{splncs04}
\bibliography{references}